\documentclass[11pt,a4paper]{article}
\usepackage[hyperref]{emnlp-ijcnlp-2019}
\usepackage{times}
\usepackage{latexsym}

\usepackage{url}
\usepackage{algorithm2e}
\aclfinalcopy 

\title{Universal Adversarial Perturbation for Text Classification}

\author{Hang Gao \\
  University of Maryland, \\ Baltimore County \\
  1000 Hilltop Cir, Baltimore, MD 21250 \\ 
  {\tt hanggao1@umbc.edu} \\\And
  Tim Oates \\
  University of Maryland, \\  Baltimore County \\
  1000 Hilltop Cir, Baltimore, MD 21250 \\
  {\tt oates@cs.umbc.edu} \\}

\date{}

\begin{document}
\maketitle
\begin{abstract}
Given a state-of-the-art deep neural network text classifier, we show the existence of a universal and very small perturbation vector (in the embedding space) that causes natural text to be misclassified with high probability. Unlike images on which a single fixed-size adversarial perturbation can be found, text is of variable
length, so we define the ``universality'' as ``token-agnostic'', where a single perturbation is applied to each token, resulting in different perturbations of flexible sizes at the sequence level. We propose an algorithm to compute universal adversarial perturbations, and show that the state-of-the-art deep neural networks are highly vulnerable to them, even though they keep the neighborhood of
tokens mostly preserved. We also show how to use these adversarial perturbations to generate adversarial text samples. The surprising existence of universal ``token-agnostic'' adversarial perturbations may reveal important properties of a text classifier.

\end{abstract}

\section{Introduction}
Deep neural networks (DNNs) are vulnerable to adversarial samples, i.e., carefully crafted samples with imperceptible perturbations designed to mislead a pre-trained model \cite{szegedy2013intriguing}, raising security concerns.

Two types of adversarial perturbations can be considered: universal or sample-dependent. Sample-dependent perturbations can vary for different samples in a dataset, while universal ones remain fixed for all samples. Adversarial samples were first studied for image DNNs with various methods proposed to generate and defend both sample-dependent perturbations \cite{szegedy2016rethinking, carlini2017towards, goodfellow2014explaining, moosavi2016deepfool, kurakin2016adversarial_e, kurakin2016adversarial_a, cisse2017houdini} and universal perturbations \cite{moosavi2017universal, mopurigb17, moosavi2017analysis}. Later, research on attacking DNNs for text applications emerged \cite{liang2017deep, ebrahimi2017hotflip, samanta2018generating, jia2017adversarial, li2018textbugger, ribeiro2018semantically}, where the definition of an ``imperceptible perturbation'' can differ from image-based perturbations. 

Although most existing adversarial attack methods on textual deep learning models are sample-dependent, recently, \newcite{behjati2019universal} proposed universal adversarial attacks on text classifiers by generating a universal sequence of words that can be added to any input to fool a classifier with high probability. Their method is effective, but is unlikely to preserve the syntax or semantic of the original input text. In this paper, we instead seek to generate universal adversarial perturbations and samples for text classifiers, by explicitly limiting the norm of changes. 

Our main contributions can be summarized as follows:

\begin{itemize}
    \item We show that a textual deep model can be vulnerable to some small token-agnostic perturbation in the token embedding space.
    \item We propose an algorithm to compute such universal adversarial perturbations. 
    \item We provide a way to generate adversarial samples in textual form, given the universal perturbations. 
\end{itemize}

\section{Related Work}
In general, attack methods on textual deep learning models can be categorized based on the following five criteria \cite{Zhang2019GeneratingTA}: (1) model access; (2) application; (3) target; (4) granularity and (5) the attacked DNNs.

\subsection{White-Box Attack}
Considering model access, white-box attacks require access to a model's full information. 

FGSM \cite{goodfellow2014explaining} was the first attack on image-based models. TextFool \cite{liang2017deep} designed three perturbation strategies: insertion, modification, and removal to generate adversarial samples, based on the concept of FGSM. \newcite{samanta2018generating}, adopting the same idea as TextFool, proposed to craft adversarial text samples by deleting or replacing the important or salient words in the text or by introducing new words to the text sample. \newcite{papernot2016crafting} used the forward derivative such as JSMA \cite{PapernotMJFCS15} to find the sequence that contributes the most towards the adversary direction. \newcite{GrossePM0M16}, on the other hand, crafted adversarial samples on input features by computing the gradient of the model Jacobian to estimate the perturbation direction. \newcite{Identify_Mengying_2018} adopted the C\&W method \cite{carlini2017towards} for attacking predictive models of medical records. Other kinds of methods are summarized in \cite{Zhang2019GeneratingTA}.

\subsection{Universal Adversarial Perturbation}
Universal adversarial perturbation was first studied for image-based deep models by \cite{moosavi2017universal}. They showed the existence of universal image-agnostic perturbations with remarkable generalization properties. Later on, \newcite{omid_generative_2017} proposed to generate universal perturbations with neural networks. \newcite{shafahi2018universal}, on the other hand, introduced an algorithm for universal adversarial training. For natural language processing, \newcite{behjati2019universal} proposed to generate universal adversarial samples by finding a universal sequence of words that can be added to any input to fool a classifier.

\section{Proposed Methods}
Given a set of samples $\mathbf{X} = \{\mathbf{x}_i, i=1,...,N\}$, and a network $f(w, \cdot)$ with frozen parameter $w$ that maps each sequence of text $\mathbf{x}_i$ onto labels, we define a universal adversarial perturbation $\delta$ to optimize the following objective:

\begin{equation}
    \max\limits_{\delta}\; \frac{1}{N}\sum_{i=1}^N l(w, \mathbf{x}_i+\delta) \;\;\mathrm{s.t.}\;\; \|\delta\|_p \le \epsilon
\end{equation}


\noindent where $l(w, \cdot)$ represents the loss used for training DNNs. We use cross-entropy as the training loss in this paper. $\epsilon$ denotes the maximum $p$-norm of $\delta$.

\begin{table*}[h!]
\centering
 \begin{tabular}{|c c c c c c c|} 
 \hline
 $\mathrm{Corpus}$ & $|\mathrm{Train}|$ & $|\mathrm{Dev}|$ & $|\mathrm{Test}|$ & $\mathrm{Task}$ & $\mathrm{Metrics}$ & $\mathrm{Domain}$\\ 
 \hline
 MRPC & 3.7k & 0.4k & 1.7k & paraphrase & acc./F1 & news \\
 \hline 
 CoLA & 8.5k & 1k  & 1k & acceptability & Matthews corr. & misc. \\ 
 \hline 
 SST-2 & 67k & 0.8k & 1.8k & sentiment & acc. & movie reviews \\ 
 \hline 
 QNLI & 105k & 5.5k & 5.4k & QA/NLI & acc. & Wikipedia \\
 \hline 
 \end{tabular}
  \caption{The statistics of the datasets used for the experiments.}
\end{table*}

\begin{algorithm}[t]
\SetKwInOut{Input}{Input}
\SetKwInOut{Output}{Output}
\SetAlgoLined
\Input{ Training samples X, perturbation maximum norm $\epsilon$, learning rate $\lambda$}
\Output{ Universal adversarial perturbation $\delta$ }
 Initialize $\delta \leftarrow 0$\;
 \For{ $\mathrm{epoch} = 1,..., N_{ep}$}{
    \For{$\mathrm{minibatch }\; B \subset X$}{
        \eIf{$\mathrm{normalize\; gradients}$}{
            $g = \nabla_{\delta}l(w, x + \delta)$ \;
            $ng = -\epsilon \frac{g}{\|g\|_p}$ \;
            $\delta \leftarrow \delta + \lambda ng$ \;
        }{
            $g = \nabla_{\delta}l(w, x + \delta)$ \;
            $\delta \leftarrow \delta + \lambda g$ \;
        }
        $\mathrm{Project} \; \delta \; \mathrm{to} \; l_p \; \mathrm{ball}$ \;
    }
 }
 \caption{The proposed algorithm to generate a universal perturbation. }
\end{algorithm}

Given a sequence of text sample $\mathbf{x}_i = x_{i, 1}, ..., x_{i, m}$ of length $m$, with $e_{i, 1}, ..., e_{i, m}$ as the embeddings for each token, the corresponding perturbed sample $\mathbf{x}_i + \delta$ is generated by adding $\delta$ to each embedding, i.e. $e_{i, 1}+\delta, ..., e_{i, m}+\delta$. The perturbation is ``universal" as ``token-agnostic", i.e. it remains the same regardless of the token to which it is applied. We present the algorithm to generate universal adversarial perturbations in Algorithm 1. 

While our definition of adversarial perturbations is similar to the one for images, the ``imperceptibleness" is different as token embeddings, unlike images, usually have no direct semantic meaning for human beings. In this paper, we define ``imperceptibleness" as the normalized intersection of the neighborhoods of a token $t$ before and after perturbation ($N^{(t)}_b$ and $N^{(t)}_a$), i.e.

\begin{equation}
    NI(t) = \frac{|N^{(t)}_b \cap N^{(t)}_a|}{\min (|N^{(t)}_a|, |N^{(t)}_b|)}
\end{equation}

\noindent and the ``imperceptibleness" score for a vocabulary $V$ is then defined as,

\begin{equation}
    NI(V) = \frac{1}{|V|} \sum_{t \in V} NI(t)
\end{equation}

\section{Experiment}
\subsection{Baseline}
Although we are interested in a single universal perturbation applicable to all tokens, we establish a strong baseline where a unique adversarial perturbation is obtained for each token in a vocabulary. The ``universality" is then considered as ``sample-agnostic'' as perturbations are now token-sensitive, but remain unchanged for the same token in different samples. These perturbations can be generated in a similar way as the one presented in Algorithm 1.

We build another set of baselines by randomizing all perturbations, and then normalizing them to the maximum $p$-norm limit $\epsilon$. 

\subsection{Classifier and Dataset}
In this paper, we focus on generating universal adversarial perturbations for text classifiers. We adopt BERT \cite{devlin2018bert} as the classifier model for its promising performance on the task of text classification. We fine tune the BERT base model with its uncased tokenizer on four GLUE datasets \cite{wang2018glue}: MRPC, CoLA, SST-2 and QNLI. The details of these datasets are shown in Table 1. For each task, we explore hyper-parameters with the following ranges, and pick the classifier with the best performance on the evaluation data. 

\begin{itemize}
    \item \textbf{Batch size:} 16, 32
    \item \textbf{Learning rate:} 5e-5, 3e-5 and 2e-5
    \item \textbf{Number of epochs:} 3, 4
\end{itemize}

We then split the training data of each task into train/dev sets and train our universal adversarial perturbation and samples on the train set with hyper-parameters picked on the dev set. We test the performance of these perturbation and samples on the original evaluation data of each task, which is never used for parameter-tuning or hyper-parameter selection for those perturbation and samples. 

\subsection{Result}

\begin{table}[ht!]
\centering
 \begin{tabular}{|c c c c c|} 
 \hline
 \multicolumn{1}{|c}{Task} & \multicolumn{4}{c|}{Maximum p-Norm $\epsilon$} \\
 \hline
 MRPC$\mathrm{(0.882)}$ & $\mathrm{0.05}$  & $\mathrm{0.1}$ & $\mathrm{0.15}$ & $\mathrm{0.2}$\\ 
 \hline
 $\mathrm{Baseline}_{\mathrm{VR}}$ & 0.882 & 0.876 & 0.876 & 0.876  \\
 $\mathrm{Baseline}_{\mathrm{SR}}$ & 0.884 & 0.884 & 0.882 & 0.884  \\
 $\mathrm{Baseline}_{\mathrm{V}}$ & 0.863 & 0.858 & 0.831 & 0.820  \\
 $\mathrm{Ours}$ & \textbf{0.860} & \textbf{0.811} & \textbf{0.770} & \textbf{0.748}  \\
 \hline 
 CoLA$\mathrm{(0.585)}$ & $\mathrm{0.05}$  & $\mathrm{0.1}$ & $\mathrm{0.15}$ & $\mathrm{0.2}$\\ 
 \hline 
 $\mathrm{Baseline}_{\mathrm{VR}}$ & 0.593 & 0.591 & 0.588 & 0.588  \\
 $\mathrm{Baseline}_{\mathrm{SR}}$ & 0.586 & 0.588 & 0.583 & 0.578  \\
 $\mathrm{Baseline}_{\mathrm{V}}$ & 0.580 & 0.575 & 0.557 & 0.346  \\
 $\mathrm{Ours}$ & \textbf{0.580} & \textbf{0.566} & \textbf{0.278} & \textbf{0.168}  \\
 \hline 
 SST-2$\mathrm{(0.929)}$ & $\mathrm{0.05}$  & $\mathrm{0.1}$ & $\mathrm{0.15}$ & $\mathrm{0.2}$\\ 
 \hline 
 $\mathrm{Baseline}_{\mathrm{VR}}$ & 0.928 & 0.925 & 0.924 & 0.924  \\
 $\mathrm{Baseline}_{\mathrm{SR}}$ & 0.928 & 0.929 & 0.930 & 0.931  \\
 $\mathrm{Baseline}_{\mathrm{V}}$ &  0.925 & 0.876 & 0.750 & 0.608 \\
 $\mathrm{Ours}$ & \textbf{0.892} & \textbf{0.823} & \textbf{0.704} & \textbf{0.585}  \\
 \hline 
 QNLI$\mathrm{(0.918)}$ & $\mathrm{0.05}$  & $\mathrm{0.1}$ & $\mathrm{0.15}$ & $\mathrm{0.2}$\\ 
 \hline 
 $\mathrm{Baseline}_{\mathrm{VR}}$ & 0.918 & 0.917 & 0.917 & 0.917  \\
 $\mathrm{Baseline}_{\mathrm{SR}}$ & 0.917 & 0.917 & 0.916 & 0.915\\
 $\mathrm{Baseline}_{\mathrm{V}}$ & 0.916 & 0.917 & 0.915 & 0.908  \\
 $\mathrm{Ours}$ & \textbf{0.916} & \textbf{0.897} & \textbf{0.844} & \textbf{0.499}  \\
 \hline 
 \end{tabular}
  \caption{The comparison of the proposed methods with baseline approaches on the evaluation data for each task. The performance of the classifier without adversarial perturbation is presented in the parentheses. $\mathrm{Baseline}_{\mathrm{VR}}$: baseline with vocabulary based random perturbations; $\mathrm{Baseline}_{\mathrm{SR}}$: baseline with a single random perturbation; $\mathrm{Baseline}_{\mathrm{V}}$: baseline with vocabulary based adversarial perturbations.}
\end{table}

\begin{table}[t]
\centering
 \begin{tabular}{|c c c c c|} 
 \hline
 \multicolumn{1}{|c}{Task} & \multicolumn{4}{c|}{Maximum p-Norm $\epsilon$} \\
 \hline
 MRPC & $\mathrm{0.05}$  & $\mathrm{0.1}$ & $\mathrm{0.15}$ & $\mathrm{0.2}$\\ 
 \hline
 $\mathrm{Baseline}_{\mathrm{VR}}$ & 0.907 & 0.830 & 0.762 & 0.697  \\
 $\mathrm{Baseline}_{\mathrm{SR}}$ & \textbf{0.974} & \textbf{0.949} & \textbf{0.919} & \textbf{0.885} \\
 $\mathrm{Baseline}_{\mathrm{V}}$ & 0.897 & 0.811 & 0.731 & 0.658  \\
 $\mathrm{Ours}$ & 0.968 & 0.935 & 0.904 & 0.863  \\
 \hline 
 CoLA & $\mathrm{0.05}$  & $\mathrm{0.1}$ & $\mathrm{0.15}$ & $\mathrm{0.2}$\\ 
 \hline 
 $\mathrm{Baseline}_{\mathrm{VR}}$ & 0.920 & 0.853 & 0.789 & 0.726  \\
 $\mathrm{Baseline}_{\mathrm{SR}}$ & \textbf{0.972} & \textbf{0.944} & \textbf{0.913} & \textbf{0.879}  \\
 $\mathrm{Baseline}_{\mathrm{V}}$ & 0.906 & 0.815 & 0.740 & 0.690  \\
 $\mathrm{Ours}$ & 0.963 & 0.931 & 0.905 & 0.863  \\
 \hline 
 SST-2 & $\mathrm{0.05}$  & $\mathrm{0.1}$ & $\mathrm{0.15}$ & $\mathrm{0.2}$\\ 
 \hline 
 $\mathrm{Baseline}_{\mathrm{VR}}$ & 0.916 & 0.840 & 0.776 & 0.719  \\
 $\mathrm{Baseline}_{\mathrm{SR}}$ & \textbf{0.975} & \textbf{0.949} & \textbf{0.920} & \textbf{0.888}  \\
 $\mathrm{Baseline}_{\mathrm{V}}$ &  0.895 & 0.804 & 0.717 & 0.654 \\
 $\mathrm{Ours}$ & 0.954 & 0.913 & 0.871 & 0.835  \\
 \hline 
 QNLI & $\mathrm{0.05}$  & $\mathrm{0.1}$ & $\mathrm{0.15}$ & $\mathrm{0.2}$\\ 
 \hline 
 $\mathrm{Baseline}_{\mathrm{VR}}$ & 0.886 & 0.793 & 0.716 & 0.649  \\
 $\mathrm{Baseline}_{\mathrm{SR}}$ & \textbf{0.975} & \textbf{0.950} & \textbf{0.922} & \textbf{0.891}\\
 $\mathrm{Baseline}_{\mathrm{V}}$ & 0.866 & 0.761 & 0.678 & 0.608 \\
 $\mathrm{Ours}$ & 0.959 & 0.943 & 0.916 & 0.877  \\
 \hline 
 \end{tabular}
  \caption{The $NI(V)$ score on each task. The number of neighbors for each token is 5 and we use cosine similarity as the distance measure between any two tokens. }
\end{table}

We use Adam \cite{kingma2014adam} as the optimization algorithm and normalized gradient for the perturbation computation. For each task, we split the training data into train/eval sets with a 0.9/0.1 ratio, train the perturbation on the train set and early stop on the eval set. The train batch size is set to 32 and the eval batch size is set to 8. $p$ is set to 2 and the learning rate $\lambda$ is set to 0.05. 

We range the maximum $p$-norm of perturbation $\epsilon$ from 0.05 to 0.2 and present the perturbed classifier's performance on the evaluation data of each task in Table 2. The choice of $\epsilon$ is determined so that the norm of the perturbation is significantly smaller than the average norm of token embeddings. For each task, we compare the proposed approach with three different baselines: $\mathrm{Baseline}_{\mathrm{VR}}$ - vocabulary based perturbations that are randomly initialized with $p$-norm scaled to $\epsilon$; $\mathrm{Baseline}_{\mathrm{SR}}$ - a single perturbation that is randomly initialized with $p$-norm scaled to $\epsilon$ and $\mathrm{Baseline}_{\mathrm{V}}$ - vocabulary based adversarial perturbations trained similarly as Algorithm 1.

The results in Table 2 indicate that (1) randomly initialized perturbations can hardly fool the classifier in spite of potentially larger $p$-norm than the trained ones; (2) our proposed single universal adversarial perturbation outperforms all the baselines on all the tasks. 

To evaluate the ``imperceptibleness" of the perturbation, we show the ``imperceptibleness" score defined in Equation (3) in Table 3. The number of neighbors for each token is set to 5 and we use cosine as the distance measure between any two tokens. Single perturbation based methods are better at preserving the neighborhood of each token than vocabulary based ones. And our proposed method is comparable with $\mathrm{Baseline}_{\mathrm{SR}}$ with slightly lower performance.

\begin{table}[t]
\centering
 \begin{tabular}{|c c c c c|} 
 \hline
 \multicolumn{5}{|c|}{Data Ratio} \\
 \hline
 $\mathrm{10\%}$ & $\mathrm{20\%}$  & $\mathrm{30\%}$ & $\mathrm{40\%}$ & $\mathrm{50\%}$\\ 
 \hline
 0.663 & 0.592 & 0.663 & 0.592 & 0.661   \\
 \hline 
 $\mathrm{60\%}$ & $\mathrm{70\%}$  & $\mathrm{80\%}$ & $\mathrm{90\%}$ & \\ 
 \hline
 0.588 & 0.589 & 0.663 & 0.662 & \\ 
 \hline
 \end{tabular}
  \caption{The evaluation result (Accuracy) on SST-2 with the universal adversarial perturbation trained on different ratios of training data. 10\% of training data is enough to generate perturbation with good performance.}
\end{table}

\begin{table}[b]
\centering
 \begin{tabular}{|p{0.85\linewidth}|c|} 
 \hline 
 Text (top: original, bottom: adversarial) & Prediction \\
 \hline
 \textbf{the} bears sniffed & 1 \\ 
 \textbf{a} bears sniffed & 0 \\
 \hline 
 i walk \textbf{and} dana runs . & 1 \\ 
 i walk \textbf{,} dana runs . & 0 \\
 \hline
  sue gave to bill \textbf{a} book . & 1\\ 
  sue gave to bill \textbf{the} book . & 0 \\
 \hline
  terry delighted \textbf{.} & 0 \\
  terry delighted \textbf{;} & 1 \\
 \hline
 dana walking \textbf{and} leslie running . & 0 \\
 dana walking \textbf{,} leslie running . & 1 \\ 
 \hline
 \end{tabular}
  \caption{A few adversarial samples generated on CoLA evaluation data. $\epsilon$ is set to 0.2.}
\end{table}

We further evaluate how much data is required to generate the universal adversarial perturbation and present the result in Table 4. The performance is reported as Accuracy on the SST-2 dataset with various ratios (from 10\% to 90\%) of training data used to generate the perturbation. The maximum $p$-norm is set to 0.15 and we adopt SGD as the optimization algorithm with momentum set to 0.9 and gradients unnormalized. The result indicates that 10\% of training data is enough to generate perturbations with comparable performance, suggesting the scalability of our method on large-scale real-world corpora. 

\subsection{Adversarial Samples}
We generate adversarial samples in textual form by (1) for each token in the vocabulary, find its closest neighbor after perturbation; (2) for each input sample, replace the token with its found neighbor such that the distance (cosine) between the two is the largest among all tokens and their corresponding neighbors in the input sample. We list some examples on CoLA in Table 5. One may argue that the low evaluation score on CoLA is be expected as the generated adversarial samples are unlikely to be linguistically acceptable. However, these results suggest that the classifier may be confused easily.

\section{Future Work}
In the future, we plan to look into  theoretical explanations for the existence of token-agnostic universal adversarial perturbations. We also seek to find a better way to utilize these universal perturbations, e.g., adversary training. 

\bibliography{emnlp-ijcnlp-2019}
\bibliographystyle{acl_natbib}

\end{document}